\documentclass[11pt,a4paper]{article}
\usepackage{acl2015}
\usepackage{times}
\usepackage{url}
\usepackage{latexsym}
\usepackage{enumitem}
\usepackage{multirow}
\usepackage{amssymb}
\usepackage{amsmath}
\usepackage{graphicx}

\title{Adapting Phrase-based Machine Translation to Normalise Medical Terms in Social Media Messages}
\author{Nut Limsopatham and Nigel Collier\\
 Department of Theoretical and Applied Linguistics \\
  University of Cambridge \\
  Cambridge, UK \\
  {\tt \{nl347,nhc30\}@cam.ac.uk}
  }
\begin{document}
\maketitle
\begin{abstract}
Previous studies have shown that health reports in social media, such as DailyStrength and Twitter, have potential for monitoring health conditions (e.g.\ adverse drug reactions, infectious diseases) in particular communities. However, in order for a machine to understand and make inferences on these health conditions, the ability to recognise when laymen's terms refer to a particular medical concept (i.e.\ text normalisation) is required. To achieve this, we propose to adapt an existing phrase-based machine translation (MT) technique and a vector representation of words to map between a social media phrase and a medical concept. We evaluate our proposed approach using a collection of phrases from tweets related to adverse drug reactions. Our experimental results show that the combination of a phrase-based MT technique and the similarity between word vector representations outperforms the baselines that apply only either of them by up to 55\%.
\end{abstract}

\vspace{3mm}
\section{Introduction}
\vspace{2mm}
Social media, such as DailyStrength\footnote{\url{http://www.dailystrength.org/}} and Twitter\footnote{\url{http://twitter.com}}, is a fast growing and potentially rich source of \emph{voice of the patient} data about experience in terms of benefits and side-effects of drugs and treatments~\cite{o2014pharmacovigilance}. However, natural language understanding from social media messages is a difficult task because of the lexical and grammatical variability of the language~\cite{baldwin2013noisy,o2014pharmacovigilance}. Indeed, language understanding by machines requires the ability to recognise when a phrase refers to a particular concept. Given a variable length phrase, an effective system should return a concept with the most similar meaning. For example, a Twitter phrase `No way I'm gettin any sleep 2nite' might be mapped to the medical concept `Insomnia' (SNOMED:193462001), when using the SNOMED-CT dictionary~\cite{spackman1997snomed}. The success of the mapping between social media phrases and formal medical concepts would enable an automatic integration between patient experiences and biomedical databases.

Existing works, e.g.~\cite{elkin2012comparison,gobbel2014development,wang2009active}, mostly focused on extracting medical concepts from medical documents. 
For example, Gobbel et al.~\shortcite{gobbel2014development} proposed a naive Bayesian-based technique to map phrases from clinical notes to medical concepts in the SNOMED-CT dictionary. Wang et al.~\shortcite{wang2009active} identified medical concepts regarding adverse drug events in electronic medical records. On the other hand, OConnor et al.~\shortcite{o2014pharmacovigilance} investigated the normalisation of medical terms in Twitter messages. In particular, they proposed to use 
the Lucene retrieval engine\footnote{\url{http://lucene.apache.org/}} to retrieve medical concepts that could be potentially mapped to a given Twitter phrase, when mapping between Twitter phrases and medical concepts.

In contrast, we argue that the medical text normalisation task~\cite{limsopatham2015ucui} can be achieved by using well-established phrase-based MT techniques, where we translate a text written in \emph{a social media language} (e.g.\ `No way I'm gettin any sleep 2nite') to a text written in \emph{a formal medical language} (e.g.\ `Insomnia'). Indeed, in this work we investigate an effective adaptation of phrase-based MT to map a Twitter phrase to a medical concept. Moreover, we propose to combine the adapted phrase-based MT technique and the similarity between word vector representations to effectively map a Twitter phrase to a medical concept.

The main contributions of this paper are three-fold:
\begin{enumerate}[nolistsep]
\item We investigate the adaptation of phrase-based MT to map a Twitter phrase to a SNOMED-CT concept.
\item We propose to combine our adaptation of phrase-based MT and the similarity between word vector representations to map Twitter phrases to formal medical concepts.
\item We thoroughly evaluate the proposed approach using phrases from our collection of tweets related to the topic of adverse drug reactions (ADRs).
\end{enumerate}

\section{Related Work}\label{sRelatedWork}

Phrase-based MT models (e.g.~\cite{koehn2003statistical,och2004alignment}) have been shown to be effective in translation between languages, as they learn local term dependencies, such as collocations, re-orderings, insertions and deletions. Koehn et al.~\shortcite{koehn2003statistical} showed that a phrase-based MT technique markedly outperformed traditional word-based MT techniques on several benchmarks. In this work, we adapt the phrase-based MT technique of Koehn et al.~\shortcite{koehn2003statistical} for the medical text normalisation task. In particular, we use the phrase-based MT technique to translate phrases from \emph{Twitter language} to \emph{formal medical language}, before mapping the translated phrases to medical concepts based on the ranked similarity of their word vector representations.

Traditional approaches for creating word vector representations treated words as atomic units~\cite{mikolov2013distributed,turian2010word}. For instance, the one-hot representation used a vector with a length of the size of the vocabulary, where one dimension is on, to represent a particular word~\cite{turian2010word}. Recently, techniques for learning high-quality word vector representations (i.e.\ distributed word representations) that could capture the semantic similarity between words, such as continuous bags of words (CBOW)~\cite{mikolov2013distributed} and global vectors (GloVe)~\cite{pennington2014glove}, have been proposed. Indeed, these distributed word representations have been effectively applied in different systems that achieve state-of-the-art performances for several NLP tasks, such as MT~\cite{mikolov2013exploiting} and named entity recognition~\cite{passos2014lexicon}. In this work, beside using word vector representations to measure the similarity between translated Twitter phrases and medical concepts, we use the similarity between word vector representations of the original Twitter phrase and a medical concept to augment the adapted phrase-based MT technique.

\section{Medical Term Normalisation}\label{sMethod}
We discuss our adaptation of phrase-based MT for medical text normalisation in Section~\ref{ssM1}. Section~\ref{ssM2} introduces our proposed approach for combining similarity score of word vector representations with the adapted phrase-based MT technique.
\subsection{Adapting Phrase-based MT}\label{ssM1}
We aim to learn a translation between a Twitter phrase (i.e.\ a phrase from a Twitter message) 
and a formal medical phrase (i.e.\ the description of a medical concept). For a given Twitter phrase $phr_t$, we find a suitable medical phrase $phr_m$ using \emph{a translation score}, based on a phrase-based model, as follows:
\begin{equation}\label{eq:tr}\small
translation_{score}(phr_m|phr_t) = p(phr_m|phr_t)
\end{equation}
where $p(phr_m|phr_t)$ can be calculated using any phrase-based MT technique, e.g.~\cite{koehn2003statistical,och2004alignment}. We then rank translated phrases $phr_m$ based on this translation score. The top-$k$ translated phrases are used for identifying the corresponding medical concept.

However, the translated phrase $phr_m$ may not be exactly matched with the description of any target medical concepts. We propose two techniques to deal with this problem. For the first technique, we rank the target concepts based on the cosine similarity between the vector representation of $phr_m$ and the vector representation of the description of each concept $desc_c$:
\begin{equation}\label{eq:mt-cos}\small
sim_{cos}(phr_m,desc_c) = \frac{V_{phr_m} \cdot V_{desc_c}} {|| V_{phr_m} ||\times|| V_{desc_c} ||}
\end{equation}
where $V_{phr_m}$ and $V_{desc_c}$ are the vector representations of $phr_m$ and $desc_c$, respectively. Any technique for creating word vector representations (e.g.\ one-hot, CBOW and GloVe) can be used. Note that if a phrase (e.g.\ $phrase_m$) contains 
several terms, we create a vector representation by summing the value of the same dimension of the vector representation of each term (i.e.\ element-wise addition).

On the other hand, the second technique also incorporates the ranked position $r$ of the translated phrase $phr_m$ when translated from the original phrase $phr_t$ using Equation~(\ref{eq:tr}). Indeed, the second technique calculates the similarity score as follows:
\begin{equation}\label{eq:mt-rcos}\small
sim_{rcos}(phr_m,desc_c) = \frac{1}{r} \cdot \frac{V_{phr_m} \cdot V_{desc_c}} {|| V_{phr_m} ||\times|| V_{desc_c} ||}
\end{equation}

\subsection{Combining Similarity Score with Phrase-based MT}\label{ssM2}
As discussed in Section~\ref{sRelatedWork}, word vector representations (e.g.\ created by CBOW or GloVe) can capture semantic similarity between words by itself. Hence, we propose to map a Twitter phrase $phr_t$ to a medical concept $c$, which is represented with a description $desc_c$, by linearly combining the cosine similarity, between vector representations of the Twitter phrase $phr_t$ and the description $desc_c$, with the similarity score computed using one of the adapted phrased-based MT techniques (introduced in Section~\ref{ssM1}), as follows:
\begin{align}\label{eq:cos}\small
sim_{combine}(phr_t,desc_c&) = \frac{V_{phr_t} \cdot V_{desc_c}} {|| V_{phr_t} ||\times|| V_{desc_c} ||} \\ \nonumber
&+ MT_a(phr_t,desc_c)
\end{align}
where $MT_a(phr_t,desc_c)$ is calculated using one of the adapted phrase-based MT techniques described in Section~\ref{ssM1}.

\section{Experimental Setup
}

\subsection{Test Collection
}
To evaluate our approach, we use a collection of 25 million tweets related to adverse drug reactions (ADRs). In particular, these tweets are related 
to cognitive enhancers~\cite{hanson2013tweaking} and anti-depressants~\cite{schneeweiss2010comparative} that can have adverse side effects. 
We use 201 ADR phrases and their corresponding SNOMED-CT concepts annotated by a PhD-level computational linguist.  
These phrases were anonymised by replacing numbers, user IDs, URIs, locations, email addresses, dates and drug names with appropriate tokens e.g.\ {\it \_NUMBER\_}.\ 

\subsection{Evaluation Approach}
We conduct experiments using 10-fold cross validation, where the Twitter phrases are randomly divided into 10 separated folds. We address this task as a ranking task, where we aim to rank the medical concept with the highest similarity score, e.g.\ calculated using Equation~(\ref{eq:mt-cos}), at the top rank. Hence, we evaluate our approach using Mean Reciprocal Rank (MRR) measure~\cite{craswell2009mean}, which is based on the the reciprocal of the rank at which the first relevant concept is viewed in the ranking. 
In addition, we compare the significant difference between the performance achieved by our proposed approach and the baselines using the paired t-test ($p<0.05$). 

\subsection{Word Vector Representation}\label{ssWVR}
We use three different techniques, including one-hot, CBOW and GloVe, to create word vector representations used in our approach (see Section~\ref{sMethod}). In particular, the vocabulary for creating the one-hot representation includes all terms in the Twitter phrases and the descriptions of the target SNOMED-CT concepts. Meanwhile, we create word vector representations based on CBOW and GloVe by using the word2vec\footnote{\url{https://code.google.com/p/word2vec/}} and GloVe\footnote{\url{http://nlp.stanford.edu/projects/glove/}} implementations. We learn the vector representations from the collections of tweets and medical articles, respectively, using window size of 10 words. The tweet collection (denoted \emph{Twitter}) contains 419,702,147 English tweets, which are related to 11 drug names and 6 cities, while the medical article collection (denoted \emph{BMC}) includes all medical articles from the BioMed Central\footnote{\url{http://www.biomedcentral.com/about/datamining}}. For both CBOW and GloVe, we create vector representations with vector sizes 50 and 200, respectively.

\begin{table*}[tb]
  \caption{MRR-5 performance of the proposed approach and the baselines. Significant differences ($p<0.05$) with the cosine similarity (\emph{vSim}) baselines with the one-hot representation, and with the corresponding distributed word representation (e.g.\ CBOW or GloVe) are denoted $^{\triangle}$ and $^{\blacktriangle}$, respectively.}
  \label{tab:1}
  \centering 
\small
\begin{tabular}{|c|c|c|c|c|c|c|c|c|c|}
\hline
\multirow{3}{*}{Approach} & \multirow{3}{*}{One-hot} & \multicolumn{4}{c|}{BMC} & \multicolumn{4}{c|}{Twitter} \\
\cline{3-10}
& & \multicolumn{2}{c|}{CBOW} & \multicolumn{2}{c|}{GloVe} & \multicolumn{2}{c|}{CBOW} & \multicolumn{2}{c|}{GloVe}\\
\cline{3-10}
& & 50 & 200 	& 50  & 200 & 50 & 200  & 50 & 200\\ \hline
vSim &	0.1675	& 0.1771	& 0.1896	& 0.1840	&	0.1869  &0.1812 &0.1813 &0.0936 &0.1807\\ \hline
bestMT &	0.2232	& 0.1926 & 0.2070 &	0.1803 &	0.2500$^{\triangle}$ & 0.2014 & 0.2047 & 0.1258 & 0.2138\\ 
top5MT &	0.2491$^{\triangle}$	& {\bf 0.1994} &	0.2104	& 0.1879 &	{\bf 0.2638}$^{\triangle\blacktriangle}$ & {\bf 0.2037} & 0.2095 & {\bf 0.1322} &0.2362 \\
top5MTr &	0.2458$^{\triangle}$	& 0.1982	&  0.2109 & 	{\bf 0.1894} &	0.2617$^{\triangle}$ & {\bf 0.2037} & {\bf 0.2096} & {\bf 0.1322} &0.2310 \\ \hline
bestMT+vSim &	0.2420$^{\triangle}$	& 0.1910 &	0.1953 &	0.1860 &	0.2532$^{\triangle}$ & 0.1891 & 0.1954 & 0.1078 & 0.2374\\
top5MT+vSim	& 0.2556$^{\triangle}$	& 0.1916 &	{\bf 0.2144} &	0.1726 &	0.2600$^{\triangle}$ & 0.1978 & 0.2068 & 0.1079 & 0.2405$^{\triangle}$\\
top5MTr+vSim	& {\bf 0.2594}$^{\triangle}$	& 0.1861 &	0.2070 &	0.1802 &	0.2590$^{\triangle}$ & 0.1959 & 0.2027 & 0.1129 & {\bf 0.2406}$^{\triangle}$\\
\hline
\end{tabular}
\end{table*}

\subsection{Learning Phrase-based Model}
We use the phrase-based MT technique of Koehn et al.~\shortcite{koehn2003statistical}, as implemented in the Moses toolkit~\cite{koehn2007moses} with default settings, to learn to translate from the Twitter language to the medical language. In particular, when training the translator, we show the learner pairs of the Twitter phrases and descriptions of the corresponding SNOMED-CT concepts.

\section{Experimental Results}\label{sResults}%\enlargethispage{\baselineskip}
We evaluate 6 different instantiations of the proposed approach discussed in Section~\ref{sMethod}, including:
\begin{enumerate}[nolistsep]
\item \emph{bestMT}: set $k=1$, when finding the translated phrase $phr_m$ for a Twitter phrase $phr_t$ (Equation~(\ref{eq:tr})), before ranking target medical concepts for the translated phrase $phr_m$ using Equation~(\ref{eq:mt-cos}).
\item \emph{top5MT}: similar to \emph{bestMT}, but set $k=5$.
\item \emph{top5MTr}: similar to \emph{top5MT}, but also consider the rank position of the translate phrases when ranking the target medical concepts by using Equation~(\ref{eq:mt-rcos}).
\item \emph{bestMT+vSim}: incorporate with the ranking generated from \emph{bestMT}, the cosine similarity between the vector representations of the Twitter phrase $phr_t$ and the description $desc_c$ of target medical concepts by using Equation~(\ref{eq:cos}). 
\item \emph{top5MT+vSim}: similar to \emph{bestMT+vSim}, but use the ranking from \emph{top5MT}.
\item \emph{top5MTr+vSim}: similar to \emph{bestMT+vSim}, but use the ranking from \emph{top5MTr}.
\end{enumerate}
Another baseline is \emph{vSim}, where we consider only the cosine similarity between the vector representations of the Twitter phrase $phr_t$ and the description $desc_c$ of target medical concepts.

Table~\ref{tab:1} compares the performance of these 6 instantiations and the \emph{vSim} baseline in terms of MRR-5. 
We firstly observe that for the \emph{vSim} baseline, excepting for word vector representation with vector size 50 learned using GloVe from the Twitter collection, word vector representations learned using either CBOW or GloVe are more effective than the one-hot representation. However, the difference between the MRR-5 performance is not statistically significant ($p>0.05$, paired t-test). In addition, word vector representations learned either using CBOW or GloVe with vector size 200 is more effective than those with vector size 50. 

Next, we find that our adaptation of phrase-based MT (i.e.\ \emph{bestMT}, \emph{top5MT} and \emph{top5MTr}) significantly ($p<0.05$) outperforms the \emph{vSim} baseline. For example, with the one-hot representation, \emph{top5MT} (MRR-5 0.2491) and \emph{top5MTr} (MRR-5 0.2458) perform significantly ($p<0.05$) better than \emph{vSim} (MRR-5 0.1675) by up to 49\%. 
Meanwhile, when using word vector representations with the vector size 200 learned using GloVe from the BMC collection, \emph{top5MT} (MRR-5 0.2638) significantly ($p<0.05$) outperforms \emph{vSim} with both the GloVe vector representation (MRR-5 0.1869) and the one-hot representation (MRR-5 0.1675). We observe the similar trends in performance when using vector representations learned from the Twitter collection. These results show that our adapted phase-based MT techniques are effective for the medical term normalisation task.

In addition, we observe the effectiveness of our combined approach (i.e.\ \emph{bestMT+vSim}, \emph{top5MT+vSim} and \emph{top5MTr+vSim}), as it further improves the performance of the adapted phrase-based MT (i.e.\ \emph{bestMT}, \emph{top5MT} and \emph{top5MTr}, respectively), when using the one-hot representation. For example, \emph{top5MTr+vSim} achieves the MRR-5 of 0.2594, while the MRR-5 of \emph{top5MTr} is 0.2458. However, the performance difference is not statistically significant. Meanwhile, when using the CBOW and GloVe vectors, the achieved performance is varied based on the collection (i.e.\ BMC or Twitter) used for learning the vectors and the size of the vectors.

\section{Conclusions}\label{sConclusions}
We have introduced our approach that adapts a phrase-based MT technique to normalise medical terms in Twitter messages. We evaluate our proposed approach using a collection of phrases from tweets related to ADRs. Our experimental results show that the proposed approach significantly outperforms an effective baseline by up to 55\%. For future work, we aim to investigate the modelling of learned vector representation, such as CBOW and GloVe, within a phrase-based MT model when normalising medical terms.

\section*{Acknowledgements}

The authors gratefully acknowledge Nestor Alvaro (Sokendai, Japan) for providing access to the Twitter/SNOMED-CT annotations which were used to derive the test collection used in these experiments. The derived dictionary and a representative sample of the word vector representations (CBOW and GloVe at 200d) are made available on Zenodo.org (DOI: http://dx.doi.org/10.5281/zenodo.27354). We wish to thank funding support from the EPSRC (grant number EP/M005089/1).

\section*{Appendix}
Tables~\ref{tab:a1} and~\ref{tab:a2} report the MRR-5 performance when using the word vector representation learned from the BMC and Twitter collections with window sizes 50, 100 and 200, using CBOW and GloVe.

\begin{table*}[b]
   \caption{\looseness-1  MRR-5 performance of the proposed approach when the word vector representation created by CBOW and GloVe is learned from the BMC collection with window sizes 50, 100 and 200. Significant differences ($p<0.05$) with the cosine similarity with the one-hot representation, and the cosine similarity with the corresponding distributed word representation vector are denoted $^{\triangle}$ and $^{\blacktriangle}$, respectively.}
  \label{tab:a1}
  \centering 
\small
\begin{tabular}{|c|c|c|c|c|c|c|c|}
\hline
\multirow{2}{*}{Approach} & \multirow{2}{*}{One-hot} & \multicolumn{3}{c|}{CBOW} & \multicolumn{3}{c|}{GloVe}\\
\cline{3-8}
& & 50 & 100 & 200 	& 50 & 100 & 200 \\ \hline
vSim &	0.1675	& 0.1771	& 0.1882 & 	0.1896	& 0.1840	& 0.1593 &	0.1869 \\ \hline
bestMT &	0.2232	& 0.1926 &	0.1956	& 0.2070 &	0.1803 &	0.2338$^{\blacktriangle}$ &	0.2500$^{\triangle}$ \\ 
top5MT &	0.2491$^{\triangle}$	& 0.1994 &	0.1971	& 0.2104	& 0.1879 &	0.2425$^{\triangle\blacktriangle}$ &	{\bf 0.2638}$^{\triangle\blacktriangle}$ \\
top5MTr &	0.2458$^{\triangle}$	& 0.1982	& 0.1971	& 0.2109 & 	0.1894 &	0.2391$^{\blacktriangle}$ &	0.2617$^{\triangle}$ \\ \hline
bestMT+vSim &	0.2420$^{\triangle}$	& 0.1910 &	0.1893 &	0.1953 &	0.1860 &	0.2375$^{\triangle\blacktriangle}$ &	0.2532$^{\triangle}$\\
top5MT+vSim	& 0.2556$^{\triangle}$	& 0.1916 &	0.2025 &	0.2144 &	0.1726 &	0.2381$^{\triangle\blacktriangle}$ &	0.2600$^{\triangle}$\\
top5MTr+vSim	& 0.2594$^{\triangle}$	& 0.1861 &	0.1918 &	0.2070 &	0.1802 &	0.2451$^{\triangle\blacktriangle}$ &	0.2590$^{\triangle}$\\
\hline
\end{tabular}
\end{table*}

\begin{table*}[H!]
  \caption{\looseness-1  MRR-5 performance of the proposed approach when the word vector representation created by CBOW and GloVe is learned from the Twitter collection with window sizes 50, 100 and 200. Significant differences ($p<0.05$) with the cosine similarity with the one-hot representation, and the cosine similarity with the corresponding distributed word representation vector are denoted $^{\triangle}$ and $^{\blacktriangle}$, respectively.}
  \label{tab:a2}
  \centering 
\small
\begin{tabular}{|c|c|c|c|c|c|c|c|}
\hline
\multirow{2}{*}{Approach} & \multirow{2}{*}{One-hot} & \multicolumn{3}{c|}{CBOW} & \multicolumn{3}{c|}{GloVe}\\
\cline{3-8}
& & 50 & 100 & 200 	& 50 & 100 & 200 \\ \hline
vSim &	0.1675 & 0.1812 &	0.1901	 & 0.1813 &	0.0936	& 0.1836 &	0.1807 \\ \hline
bestMT & 0.2232 &	0.2014 &	0.1993 &	0.2047 &	0.1258 &	0.2114 &	0.2138 \\
top5MT & 0.2491$^{\triangle}$ &	0.2037 &	0.2060	& 0.2095 &	0.1322 &	0.2320	& 0.2362 \\
top5MTr & 0.2458$^{\triangle}$ &	0.2037 &	0.2037 &	0.2096 &	0.1322 &	0.2279 &	0.2310 \\ \hline
bestMT+vSim & 0.2420$^{\triangle}$ &	0.1891	& 0.1959 &	0.1954	& 0.1078 &	0.2161 &	0.2374 \\
top5MT+vSim & 0.2556$^{\triangle}$ &	0.1978	& 0.2033 &	0.2068	& 0.1079 &	0.2420$^{\triangle}$ &	0.2405$^{\triangle}$ \\
top5MTr+vSim & {\bf 0.2594}$^{\triangle}$ &	0.1959	& 0.1913 &	0.2027	& 0.1129 &	0.2352	& 0.2406$^{\triangle}$ \\
\hline
\end{tabular}
\end{table*}

\end{document}